\documentclass[11pt,a4paper]{article}
\usepackage[hyperref]{acl2019}
\usepackage{times}
\usepackage{latexsym}
\usepackage{url}

\usepackage{latexsym}
\usepackage{amsmath}
\DeclareMathOperator{\sign}{sign}

\newcommand{\tf}[1]{\textbf{#1}}
\usepackage{amssymb,amsmath,amsthm,enumitem}
\usepackage{comment}

\usepackage{url}
\usepackage{graphics}

\usepackage{times}
\usepackage{graphicx}
\usepackage{amsmath,amssymb,enumerate,bbm,color,alltt}
\usepackage{graphicx}
\usepackage{float}
\usepackage{stfloats}
\usepackage{subfig}
\usepackage{longtable}
\usepackage{tabularx}
\usepackage{multirow}
\usepackage{booktabs}
\usepackage{subfig}
\usepackage{url}

\usepackage{hyperref}
\usepackage{url}
\usepackage{tabularx}
\newcolumntype{C}[1]{>{\centering\arraybackslash}p{#1}}
\usepackage{wrapfig,lipsum,booktabs}
\usepackage{amsmath}
\usepackage{amssymb}
\usepackage{subfig}
\usepackage{adjustbox} 
\usepackage{caption}

\usepackage{amsmath,amsthm,amsfonts,amssymb}
\usepackage{bm}
\usepackage{algorithm, algorithmic}
\usepackage{graphicx}
\usepackage{sidecap}
\usepackage{mathrsfs}
\usepackage{multirow, multicol}

\newcommand{\bbR}{\mathbb{R}}

\def\1{\bm{1}}

\DeclareMathAlphabet{\mathsfit}{\encodingdefault}{\sfdefault}{m}{sl}
\SetMathAlphabet{\mathsfit}{bold}{\encodingdefault}{\sfdefault}{bx}{n}

\aclfinalcopy 

\title{Learning Compressed Sentence Representations\\for On-Device Text Processing}

\author{Dinghan Shen$^{\mathbf{1*}}$, ~Pengyu Cheng$^{\mathbf{1}}\thanks{~~~Equal contribution.}~~$, ~Dhanasekar Sundararaman$^{\mathbf{1}}$
	\smallskip 
	~  \\
	\bf{Xinyuan Zhang$^{\mathbf{1}}$, Qian Yang$^{\mathbf{1}}$, ~Meng Tang$^{\mathbf{3}}$, ~Asli Celikyilmaz$^{\mathbf{2}}$, ~Lawrence Carin$^{\mathbf{1}}$} \\
	\smallskip 
	$^{\mathbf{1}}$ Duke University~~~~
	$^{\mathbf{2}}$ Microsoft Research~~~
	$^{\mathbf{3}}$ Stanford University~~~~ \\
	{\tt dinghan.shen@duke.edu} 
}

\date{}

\begin{document}
\maketitle
\begin{abstract}
Vector representations of sentences, trained on massive text corpora, are widely used as generic sentence embeddings across a variety of NLP problems. 
The learned representations are generally assumed to be \emph{continuous} and \emph{real-valued}, giving rise to a large memory footprint and slow retrieval speed, which hinders their applicability to low-resource (memory and computation) platforms, such as mobile devices. 
In this paper, we propose four different strategies to transform \emph{continuous} and generic sentence embeddings into a \emph{binarized} form, while preserving their rich semantic information. 
The introduced methods are evaluated across a wide range of downstream tasks,
where the binarized sentence embeddings are demonstrated to degrade performance by only about $2\%$ relative to their continuous counterparts, while reducing the storage requirement by over $98\%$.
Moreover, with the learned binary representations, the semantic relatedness of two sentences can be evaluated by simply calculating their Hamming distance, which is more computational efficient compared with the inner product operation between continuous embeddings. 
Detailed analysis and case study further validate the effectiveness of proposed methods.
\end{abstract}

\section{Introduction}
Learning general-purpose sentence representations from large training corpora has received widespread attention in recent years. The learned sentence embeddings can encapsulate rich prior knowledge of natural language, which has been demonstrated to facilitate a variety of downstream tasks (\emph{without} fine-tuning the encoder weights). The generic sentence embeddings can be trained either in an unsupervised manner \citep{Kiros2015SkipThoughtV, Hill2016LearningDR, Jernite2017DiscourseBasedOF, Gan2017LearningGS, Logeswaran2018AnEF, Pagliardini2018UnsupervisedLO}, or with supervised tasks such as paraphrase identification \citep{Wieting2016TowardsUP}, natural language inference \citep{Conneau2017SupervisedLO}, discourse
relation classification \citep{Nie2017DisSentSR}, machine translation \citep{Wieting2018ParaNMT50MPT}, \emph{etc}.

Significant effort has been devoted to designing better training objectives for learning sentence embeddings. However, prior methods typically assume that the general-purpose sentence representations are \emph{continuous} and \emph{real-valued}. However, this assumption is sub-optimal from the following perspectives: \emph{\romannumeral1}) the sentence embeddings require large storage or memory footprint; \emph{\romannumeral2}) it is computationally expensive to retrieve semantically-similar sentences, since every sentence representation in the database needs to be compared, and the inner product operation is computationally involved.
These two disadvantages hinder the applicability of generic sentence representations to mobile devices, where a relatively tiny memory footprint and low computational capacity are typically available \citep{ravi2018self}.

In this paper, we aim to mitigate the above issues by binarizing the continuous sentence embeddings. Consequently, the embeddings require much smaller footprint, and similar sentences can be obtained by simply selecting those with closest binary codes in the Hamming space \citep{kiros2018inferlite}. One simple idea is to naively binarize the continuous vectors by setting a hard threshold. However, we find that this strategy leads to significant performance drop in the empirical results. Besides, the dimension of the binary sentence embeddings cannot be flexibly chosen with this strategy, further limiting the practice use of the direct binarization method.

In this regard, we propose three alternative strategies to parametrize the transformation from pre-trained generic continuous embeddings to their binary forms. Our exploration spans from \emph{simple} operations, such as a random projection, to \emph{deep} neural network models, such as a \emph{regularized} autoencoder. Particularly, we introduce a semantic-preserving objective, which is augmented with the standard autoenoder architecture to encourage abstracting informative binary codes. InferSent \citep{Conneau2017SupervisedLO} is employed as the testbed sentence embeddings in our experiments, but the binarization schemes proposed here can easily be extended to other pre-trained general-purpose sentence embeddings. We evaluate the quality of the learned general-purpose binary representations using the SentEval toolkit \citep{Conneau2017SupervisedLO}. It is observed that the inferred binary codes successfully maintain the semantic features contained in the continuous embeddings, and only lead to around $2\%$ performance drop on a set of downstream NLP tasks, while requiring merely $1.5\%$ memory footprint of their continuous counterparts. 

Moreover, on several sentence matching benchmarks, we demonstrate that the relatedness between a sentence pair can be evaluated by simply calculating the Hamming distance between their binary codes, which perform on par with or even superior than measuring the cosine similarity between continuous embeddings (see Table~\ref{tab:eval}).
Note that computing the Hamming distance is much more computationally efficient than the inner product operation in a continuous space. 
We further perform a $K$-nearest neighbor sentence retrieval experiment on the SNLI dataset \citep{Bowman2015ALA}, and show that those semantically-similar sentences can indeed be efficiently retrieved with off-the-shelf binary sentence representations.
Summarizing, our contributions in this paper are as follows: 

\emph{\romannumeral1}) to the best of our knowledge, we conduct the first systematic exploration on learning general-purpose \emph{binarized} (memory-efficient) sentence representations, and four different strategies are proposed; 

\emph{\romannumeral2}) an autoencoder architecture with a carefully-designed \emph{semantic-preserving loss} exhibits strong empirical results on a set of downstream NLP tasks; 

\emph{\romannumeral3}) more importantly, we demonstrate, on several sentence-matching datasets, that simply evaluating the \emph{Hamming distance} over binary representations performs on par or even better than calculating the \emph{cosine similarity} between their continuous counterparts (which is less computationally-efficient). 
\vspace{-2mm}
\section{Related Work}
Sentence representations pre-trained from a large amount of data have been shown to be effective when transferred to a wide range of downstream tasks. Prior work along this line can be roughly divided into two categories: \emph{\romannumeral1}) pre-trained models that require fine-tuning on the specific transferring task \citep{dai2015semi, Ruder2018UniversalLM, radford2018improving, devlin2018bert, Cer2018UniversalSE}; \emph{\romannumeral2}) methods that extract general-purpose sentence embeddings, which can be effectively applied to downstream NLP tasks \emph{without} fine-tuning the encoder parameters \citep{Kiros2015SkipThoughtV, Hill2016LearningDR, Jernite2017DiscourseBasedOF, Gan2017LearningGS, Adi2017FinegrainedAO, Logeswaran2018AnEF, Pagliardini2018UnsupervisedLO, tang2018improving}. Our proposed methods belong to the second category and provide a generic and easy-to-use encoder to extract highly informative sentence representations. However, our work is unique since the embeddings inferred from our models are binarized and compact, and thus possess the advantages of small memory footprint and much faster sentence retrieval.

Learning memory-efficient embeddings with deep neural networks has attracted substantial attention recently. One general strategy towards this goal is to extract discrete or binary data representations \citep{jang2016categorical, shu2017compressing, dai2017stochastic, chen2018learning, Shen2018NASHTE, Tissier2018NearlosslessBO}. Binarized embeddings are especially attractive because they are more memory-efficient (relative to discrete embeddings), and they also enjoy the advantages of fast retrieval based upon a Hamming distance calculation. Previous work along this line in NLP has mainly focused on learning compact representations at the word-level \citep{shu2017compressing, chen2018learning, Tissier2018NearlosslessBO}, while much less effort has been devoted to extracting binarized embeddings at the sentence-level. Our work aims to bridge this gap, and serves as an initial attempt to facilitate the deployment of state-of-the-art sentence embeddings on on-device mobile applications. 

Our work is also related to prior research on semantic hashing, which aims to learn binary text embeddings specifically for the information retrieval task \citep{salakhutdinov2009semantic, zhang2010self, wang2014hashing, xu2015convolutional, Shen2018NASHTE}. However, these methods are typically trained and evaluated on documents that belong to a specific domain, and thus cannot serve as generic binary sentence representation applicable to a wide variety of NLP taks. In contrast, our model is trained on large corpora and seeks to provide general-purpose binary representations that can be leveraged for various application scenarios. 

\section{Proposed Approach}
We aim to produce compact and binarized representations from continuous sentence embeddings, and preserve the associated semantic information. Let $x$ and $f$ denote, respectively, an input sentence and the function defined by a pre-trained general-purpose sentence encoder. Thus, $f(x)$ represents the continuous embeddings extracted by the encoder. The goal of our model is to learn a universal transformation $g$ that can convert $f(x)$ to highly informative binary sentence representations, \emph{i.e.}, $g(f(x))$, which can be used as generic features for a collection of downstream tasks. We explore four strategies to parametrize the transformation $g$. 

\subsection{Hard Threshold}
We use ${h}$ and ${b}$ to denote the continuous and binary sentence embeddings, respectively, and $L$ denotes the dimension of $h$. The first method to binarize the continuous representations is to simply convert each dimension to either $0$ or $1$ based on a hard threshold. This strategy requires no training and directly operates on pre-trained continuous embeddings. Suppose $s$ is the hard threshold, we
have, for $i = 1, 2, ......, L$:
\begin{align}
b^{(i)} = \boldsymbol{1}_{h^{(i)} > s} = \frac{{\sign}(h^{(i)}-s) + 1}{2}, \label{eq:direct} 
\end{align}
One potential issue of this direct binarization method is that the information contained in the continuous representations may be largely lost, since there is no training objective encouraging the preservation of semantic information in the produced binary codes \citep{Shen2018NASHTE}. 
Another disadvantage is that the length of the resulting binary code must be the same as the original continuous representation, and can not be flexibly chosen. In practice, however, we may want to learn shorter binary embeddings to save more memory footprint or computation. 
\subsection{Random Projection}
To tackle the limitation of the above direct binarization method, we consider an alternative strategy that requires no training either: simply applying a random projection over the pre-trained continuous representations. \citet{Wieting2018NoTR} has shown that random sentence encoders can effectively construct universal sentence embeddings from word vectors, while possessing the flexibility of adaptively altering the embedding dimensions. 
Here, we are interested in exploring whether a random projection would also work well while transforming continuous sentence representations into their binary counterparts.

We randomly initialize a matrix ${W} \in \mathbb{R}^{D \times L}$, where $D$ denotes the dimension of the resulting binary representations. Inspired by the standard initialization heuristic employed in \citep{glorot2010understanding, Wieting2018NoTR}, the values of the matrix are initialized as sampled uniformly. For $i = 1, 2,\dots, D$ and $j = 1, 2, \dots, L$, we have:
\begin{align}
W_{i, j} \sim \text{Uniform}(- \frac{1}{\sqrt{D}}, \frac{1}{\sqrt{D}}), \label{eq:init} 
\end{align}
After converting the continuous sentence embeddings to the desired dimension $D$ with the matrix randomly initialized above, we further apply the operation in \eqref{eq:direct} to binarize it into the discrete/compact form. The dimension $D$ can be set arbitrarily with this approach, which is easily applicable to any pre-trained sentence embeddings (since no training is needed). This strategy is related to the Locality-Sensitive Hashing (LSH) for inferring binary embeddings \citep{van2010online}. 
\begin{figure*}
    \centering
    \includegraphics[width=0.76\textwidth]{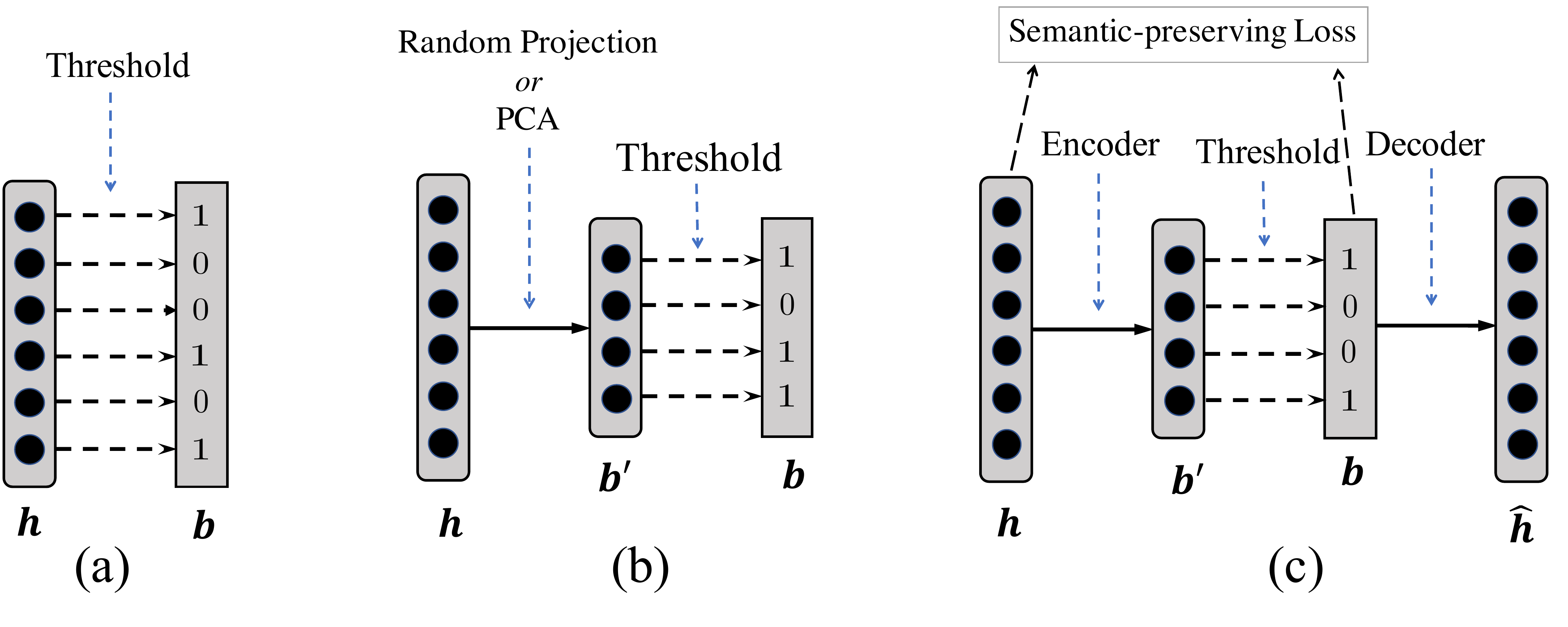}
    \vspace{-3mm}
    \caption{Proposed model architectures: (a) direct binarization with a hard threshold $s$; (b) reducing the dimensionality with either a random projection or PCA, followed by a binarization step; (c) an encoding-decoding framework with an additional semantic-preserving loss. }
    \label{fig:model}
    \vspace{-5mm}
\end{figure*}

\subsection{Principal Component Analysis}
We also consider an alternative strategy to adaptively choose the dimension of the resulting binary representations. Specifically, Principal Component Analysis (PCA) is utilized to reduce the dimensionality of pre-trained continuous embeddings. 

Given a set of sentences $\{x_i\}_{i=1}^N$ and their corresponding continuous embeddings $\{ {h}_i \}_{i=1}^N \subset \bbR^L$, we learn a projection matrix to reduce the embedding dimensions while keeping the embeddings distinct as much as possible.
After centralizing the embeddings as $h_i = h_i - \frac{1}{N}\sum_{i=1}^N h_i$, the data, as a matrix $H = (h_1,h_2,\dots,h_N)$, has the singular value decomposition (SVD):
\begin{equation*}
    H= U\Lambda V^T,
\end{equation*}
where ${\Lambda}$ is an $L \times N $ matrix with descending singular values of $ X$ on its diagonal, with $U$ and $V$ orthogonal matrices.
Then the correlation matrix can be written as: ${H}{H}^T = {U} {\Lambda}^2 {U}^T$. Assume that the diagonal matrix ${\Lambda}^2 = \text{diag}(\lambda_1,\lambda_2,\dots,\lambda_L)$ has descending elements  $\lambda_1 \geq \lambda_2 \geq \dots \geq \lambda_L \geq 0$.
We select first $D$ rows of ${U}$ as our projection matrix ${W} = {U}_{1:D}$, then the correlation matrix of ${WH}$ is ${WH}{H}^T{W}^T = \text{diag}(\lambda_1,\lambda_2,\dots,\lambda_D)$, which indicates that the embeddings are projected to $D$ independent and most distinctive axes. 

After projecting continuous embeddings to a representative lower dimensional space, we apply the hard threshold function at the position $0$ to obtain the binary representations (since the embeddings are zero-centered).
\subsection{Autoencoder Architecture}
\label{sec:ae}
The methods proposed above suffer from the common issue that the model objective does not explicitly encourage the learned binary codes to retain the semantic information of the original continuous embeddings, and a separate binarization step is employed after training. To address this shortcoming, we further consider an autoencoder architecture, that leverages the reconstruction loss to hopefully endow the learned binary representations with more informatio\label{sec:method}n. Specifically, an encoder network is utilized to transform the continuous into a binary latent vector, which is then reconstructed back with a decoder network.

For the encoder network, we use a matrix operation, followed by a binarization step, to extract useful features (similar to the random projection setup). Thus, for $i = 1, 2, \dots, D$, we have:
\begin{align}
b^{(i)} & = \boldsymbol{1}_{\sigma(W_{i} \cdot h + k^{(i)}) > s^{(i)}} \nonumber  \\
\vspace{2mm}
& = \frac{{\sign}(\sigma(W_{i}\cdot h + k^{(i)}) -s^{(i)}) + 1}{2}, \label{eq:enc}
\end{align}
where $k$ is the bias term and $k^{(i)}$ corresponds to the $i$-th element of $k$. $s^{(i)}$ denotes the threshold determining whether the $i$-th bit is $0$ or $1$. During training, we may use either \emph{deterministic} or \emph{stochastic} binarization upon the latent variable. For the deterministic case, $s^{(i)} = 0.5$ for all dimensions; in the stochastic case, $s^{(i)}$ is uniformly sampled as: $s^{(i)}\sim{\rm Uniform}(0, 1)$. We conduct an empirical comparison between these two binarization strategies in Section~\ref{sec:exp}.

Prior work have shown that linear decoders are favorable for learning binary codes under the encoder-decoder framework \citep{carreira2015hashing, dai2017stochastic, Shen2018NASHTE}. Inspired by these results, 
we employ a linear transformation to reconstruct the original continuous embeddings from the binary codes:
\begin{align}
\vspace{-2mm}
\hat{h}^{(i)} = W_{i}^{\prime} \cdot b + {k ^{\prime}}^{(i)}, \label{eq:decoder}
\vspace{-2mm}
\end{align}
where $W^\prime$ and $k^\prime$ are weight and bias term respectively, which are learned.  
The mean square error between $h$ and $\hat{h}$ is employed as the reconstruction loss:
\begin{align}
\vspace{-2mm}
\mathcal{L}_{\textit{rec}} = \frac{1}{D} \sum_{i=1}^{D}(h^{(i)} - \hat{h}^{(i)})^2, \label{eq:rec}
\vspace{-2mm}
\end{align}
This objective imposes the binary vector $b$ to encode more information from $h$ (leading to smaller reconstruction error). Straight-through (ST) estimator \citep{hinton2012neural} is utilized to estimate the gradients for the binary variable. The autoencoder model is optimized by minimizing the reconstruction loss for all sentences. After training, the encoder network is leveraged as the transformation to convert the pre-trained continuous embeddings into the binary form. 
\subsubsection{Semantic-preserving Regularizer}
Although the reconstruction objective can help the binary variable to endow with richer semantics, there is no loss that explicitly encourages the binary vectors to preserve the similarity information contained in the original continuous embeddings. Consequently, the model may lead to small reconstruction error but yield sub-optimal binary representations \citep{Tissier2018NearlosslessBO}. 
To improve the semantic-preserving property of the inferred binary embeddings, we introduce an additional objective term. 

Consider a triple group of sentences $(x_\alpha, x_\beta, x_\gamma)$, whose continuous embeddings are $(h_\alpha, h_\beta, h_\gamma)$, respectively. Suppose that the cosine similarity between $h_\alpha$ and $h_\beta$ is larger than that between $h_\beta$ and $h_\gamma$, then it is desirable that the Hamming distance between $b_\alpha$ and $b_\beta$ should be smaller than that between $b_\beta$ and $b_\gamma$ (notably, both large cosine similarity and small Hamming distance indicate that two sentences are semantically-similar).

Let $d_c(\cdot,\cdot)$ and $d_h(\cdot,\cdot)$ denote the cosine similarity and Hamming distance (in the continuous and binary embedding space), respectively. Define $l_{\alpha,\beta,\gamma}$ as an indicator such that, $l_{\alpha,\beta,\gamma} = 1$ if $d_c(h_\alpha,h_\beta) \geq d_c(h_\beta,h_\gamma)$, and $l_{\alpha, \beta, \gamma} = -1$ otherwise. The semantic-preserving regularizer is then defined as:
\begin{align}
    \mathcal{L}_{\textit{sp}} = \sum_{\alpha, \beta, \gamma} \max\{0, l_{\alpha, \beta, \gamma} [d_h(b_\alpha, b_\beta)  - d_h( b_\beta, b_\gamma)]\},
\end{align}
By penalizing $L_{sp}$, the learned transformation function $g$ is explicitly encouraged to retain the semantic similarity information of the original continuous embeddings. Thus, the entire objective function to be optimized is:
\begin{align}
    \mathcal{L}  = \mathcal{L}_{\textit{rec}} + \lambda_{sp} \mathcal{L}_{\textit{sp}}, \label{eq:sp}
\end{align}
where $\lambda_{sp}$ controls the relative weight between the reconstruction loss ($\mathcal{L}_{\textit{rec}}$) and semantic-preserving loss ($\mathcal{L}_{\textit{sp}}$).

\subsection{Discussion}
Another possible strategy is to directly train the general-purpose binary embeddings from scratch, \emph{i.e.}, jointly optimizing the continuous embeddings training objective and continuous-to-binary parameterization. However, our initial attempts demonstrate that this strategy leads to inferior empirical results. This observation is consistent with the results reported in \citep{kiros2018inferlite}. Specifically, a binarization layer is directly appended over the InferSent architecture \citep{Conneau2017SupervisedLO} during training, which gives rise to much larger drop in terms of the embeddings' quality (we have conducted empirical comparisons with \citep{kiros2018inferlite} in Table~\ref{tab:eval}). Therefore, here we focus on learning universal binary embeddings based on pretained continuous sentence representations. 


\begin{table*}[ht]
  \centering
  \begin{small}
  \vspace{-4mm}
  \setlength{\tabcolsep}{4pt}
  \def\arraystretch{1.15}
    \begin{tabular}{c||c|c|c|c|c|c|c|c|c|c}
    \toprule[1.2pt]
    \tf{Model} & \tf{Dim} & \tf{MR}    & \tf{CR} & \tf{SUBJ}  & \tf{MPQA} & \tf{SST}  & \tf{STS14} & \tf{STSB} & \tf{SICK-R} & \tf{MRPC} \\
    \hline
    \multicolumn{11}{c}{\emph{\textbf{Continuous} (\textbf{dense}) sentence embeddings}} \\
    \hline
    fastText-BoV  & 300 & 78.2  & 80.2  & 91.8  & 88.0  & 82.3 & .65/.63 & 58.1/59.0& 0.698 & 67.9/74.3 \\
    SkipThought  & 4800 & 76.5  & 80.1 & 93.6  & 87.1  & 82.0 & .29/.35 &41.0/41.7  & 0.595 & 57.9/66.6 \\
    SkipThought-LN & 4800 & 79.4  & 83.1  & \textbf{93.7}  & 89.3  & 82.9 & .44/.45 & - & - & - \\
    \hline
    InferSent-FF & 4096 & 79.7  & 84.2  & 92.7  & 89.4  & 84.3 & .68/.66 &55.6/56.2 & 0.612 & \textbf{67.9/73.8} \\
    InferSent-G & 4096 & \textbf{81.1}  & \textbf{86.3}  & 92.4  & \textbf{90.2} & \textbf{84.6} & \textbf{.68/.65} & \textbf{70.0/68.0 }&\textbf{0.719} & {67.4/73.2} \\
    \hline 
    \multicolumn{11}{c}{\emph{\textbf{Binary} (\textbf{compact}) sentence embeddings}} \\
    \hline 
    InferLite-short & 256 & 73.7 & 81.2 & 83.2 & 86.2 & 78.4 & 0.61/- &63.4/63.3 & 0.597& 61.7/70.1\\
    InferLite-medium & 1024 & 76.3 & 83.2 & 87.8 & 88.4 & 81.3 & 0.67/- & 64.9/64.9  & 0.642 & 64.1/72.0\\
    InferLite-long & 4096 & 77.7 & 83.7 & 89.6 & 89.1 & 82.3 & 0.68/- & 67.9/67.6 & 0.663 & 65.4/\textbf{72.9} \\
    \hline
    HT-binary & 4096 & 76.6 & 79.9 & \textbf{91.0} & 88.4 & 80.6 & .62/.60 & 55.8/53.6 & 0.652 & 65.6/70.4 \\
    Rand-binary & 2048 & 78.7  & 82.7  & 90.4 & 88.9 & 81.3   & .66/.63 & 65.1/62.3 & {0.704} & 65.7/70.8 \\
    PCA-binary  & 2048 & 78.4 & 84.5 & 90.7 & 89.4 & 81.0 & .66/.65 & 63.7/62.8 & 0.518 & 65.0/ 69.7 \\
    AE-binary &  2048 & 78.7 & \textbf{84.9} & 90.6 & 89.6 & 82.1  & .68/.66 & {71.7/69.7} & {0.673} & 65.8/70.8 \\
    AE-binary-SP & 2048 & \textbf{79.1} & {84.6} & 90.8  & \textbf{90.0}  & \textbf{82.7} & \textbf{.69/.67} & \textbf{73.2/70.6} & \textbf{0.705} & \textbf{67.2}/{72.0} \\
    \bottomrule[1.2pt]
    \end{tabular}%
\caption{Performance on the test set for $10$ downstream tasks. The STS14, STSB and MRPC are evaluated with Pearson and Spearman correlations, and SICK-R is measured with Pearson correlation. All other datasets are evaluated with test accuracy. InferSent-G uses Glove (G) as the word embeddings, while InferSent-FF employs FastText (F) embeddings with Fixed (F) padding. The empirical results of InferLite with different lengths of binary embeddings, \emph{i.e.}, 256, 1024 and 4096, are considered. }
  \label{tab:eval}%
  \end{small}
  \vspace{-4mm}
\end{table*}%

\section{Experimental setup} \label{sec:exp}
\subsection{Pre-trained Continuous Embeddings}
Our proposed model aims to produce highly informative binary sentence embeddings based upon pre-trained continuous representations. In this paper, we utilize InferSent \citep{Conneau2017SupervisedLO} as the continuous embeddings (given its effectiveness and widespread use). Note that all four proposed strategies can be easily extended to other pre-trained general-purpose sentence embeddings as well. 

Specifically, a bidirectional LSTM architecture along with a max-pooling operation over the hidden units is employed as the sentence encoder, and the model parameters are optimized on the natural language inference tasks, \emph{i.e.}, Standford Natural Language Inference (SNLI) \citep{Bowman2015ALA} and Multi-Genre Natural Language Inference (MultiNLI) datasets \citep{williams2017broad}.  

\vspace{-1mm}
\subsection{Training Details}
Our model is trained using Adam \citep{kingma2014adam}, with a value $1 \times 10^{-5}$ as the learning rate for all the parameters. The number of bits (\emph{i.e.}, dimension) of the binary representation is set as 512, 1024, 2048 or 4096, and the best choice for each model is chosen on the validation set, and the corresponding test results are presented in Table~\ref{tab:eval}. The batch size is chosen as $64$ for all model variants. The hyperparameter over $\lambda_{\textit{sp}}$ is selected from $\{0.2, 0.5, 0.8, 1\}$ on the validation set, and $0.8$ is found to deliver the best empirical results. The training with the autoencoder setup takes only about $1$ hour to converge, and thus can be readily applicable to even larger datasets.  
\vspace{-1mm}
\subsection{Evaluation}
To facilitate comparisons with other baseline methods, we use SentEval toolkit\footnote{https://github.com/facebookresearch/SentEval} \citep{conneau2018senteval} to evaluate the learned binary (compact) sentence embeddings.  Concretely, the learned representations are tested on a series of downstream tasks to assess their transferability (with the encoder weights fixed), which can be categorized as follows:   
\vspace{-1mm}
\begin{itemize}
    \item \textbf{Sentence classification}, including sentiment analysis (MR, SST),  product reviews (CR), subjectivity classification (SUBJ), opinion polarity detection (MPQA) and question type classification (TREC). A linear classifier is trained with the generic sentence embeddings as the input features. The default SentEval settings is used for all the datasets.
    \item {\textbf{Sentence matching}}, which comprises semantic relatedness (SICK-R, STS14, STSB) and paraphrase detection (MRPC). Particularly, each  pair of sentences in STS14 dataset is associated with a similarity score from $0$ to $5$ (as the corresponding label). Hamming distance between the binary representations is directly leveraged as the prediction score (\emph{without} any classifier parameters). 
\end{itemize}
\vspace{-1mm}
For the sentence matching benchmarks, to allow fair comparison with the continuous embeddings, we do not use the same classifier architecture in SentEval. Instead, we obtain the predicted relatedness by directly computing the \emph{cosine similarity} between the continuous embeddings. Consequently, there are no classifier parameters for both the binary and continuous representations. The same valuation metrics in SentEval\citep{conneau2018senteval} are utilized for all the tasks. For MRPC, the predictions are made by simply judging whether a sentence pair's score is larger or smaller than the averaged Hamming distance (or cosine similarity).
\vspace{-1mm}
\subsection{Baselines}
\vspace{-1mm}
We consider several strong baselines to compare with the proposed methods, which include both \tf{continuous} (dense) and \tf{binary} (compact) representations. For the continuous generic sentence embeddings, we make comparisons with fastText-BoV \citep{joulin2016bag}, Skip-Thought Vectors \citep{Kiros2015SkipThoughtV} and InferSent \citep{Conneau2017SupervisedLO}. As to the binary embeddings, we consider the binarized version of InferLite \citep{kiros2018inferlite}, which, as far as we are concerned, is the only general-purpose binary representations baseline reported.
\section{Experimental Results}
We experimented with five model variants to learn general-purpose binary embeddings: \emph{HT-binary} (hard threshold, which is selected from $\{0, 0.01, 0.1\}$ on the validation set), \emph{Rand-binary} (random projection), \emph{PCA-binary} (reduce the dimensionality with principal component analysis), \emph{AE-binary} (autoencoder with the reconstruction objective) and \emph{AE-binary-SP} (autoencoder with both the reconstruction objective and Semantic-Preserving loss). Our code will be released to encourage future research.
\subsection{Task transfer evaluation}
We evalaute the binary sentence representations produced by different methods with a set of transferring tasks. The results are shown in Table~\ref{tab:eval}. The proposed autoencoder architecture generally demonstrates the best results. Especially while combined with the semantic-preserving loss defined in \eqref{eq:sp}, AE-binary-SP exhibits higher performance compared with a standard autoencoder. It is worth noting that the Rand-binary and PCA-binary model variants also show competitive performance despite their simplicity. These strategies are also quite promising given that no training is required given the pre-trained continuous sentence representations.

Another important result is that, the AE-binary-SP achieves competitive results relative to the InferSent, leading to only about $2\%$ loss on most datasets and even performing at par with InferSent on several datasets, such as the MPQA and STS14 datasets. 
On the sentence matching tasks, the yielded binary codes are evaluated by merely utilizing the hamming distance features (as mentioned above).
To allow fair comparison, we compare the predicted scores with the cosine similarity scores based upon the continuous representations (there are no additional parameters for the classifier).
The binary codes brings out promising empirical results relative to their continuous counterparts, and even slightly outperform InferSent on the STS14 dataset. 

\begin{table*}
  \centering
  \begin{scriptsize}
  \vspace{-3mm}
  \def\arraystretch{1.0}
    \begin{tabular}{p{2.6in}| p{2.6in}}
    \toprule[1.2pt]
    \tf{Hamming Distance (binary embeddings)} & \tf{Cosine Similarity (continuous embeddings)}  \\
    \hline 
    \multicolumn{2}{c}{\textbf{Query: \textcolor{red}{Several} people are sitting in a \textcolor{blue}{movie theater} .}} \\
    \hline
    \textcolor{red}{A group of} people watching a \textcolor{blue}{movie} at a \textcolor{blue}{theater} . & \textcolor{red}{A group of} people watching a movie at a theater . \\
    \textcolor{red}{A crowd of people} are watching a \textcolor{blue}{movie} indoors . &   A man is watching a \textcolor{blue}{movie} in a \textcolor{blue}{theater} . \\
    A man is watching a \textcolor{blue}{movie} in a \textcolor{blue}{theater} . & \textcolor{red}{Some} people are sleeping on a sofa in front of the television .  \\
    \midrule
    
    \multicolumn{2}{c}{\textbf{Query: A \textcolor{red}{woman} crossing a \textcolor{blue}{busy} downtown street .}} \\
    \hline
    A  \textcolor{red}{lady} is walking down a \textcolor{blue}{busy} street . & A  \textcolor{red}{woman} walking on the street downtown .    \\
    A  \textcolor{red}{woman} is on a \textcolor{blue}{crowded} street . & A  \textcolor{red}{lady} is walking down a \textcolor{blue}{busy} street .  \\
    A  \textcolor{red}{woman} walking on the street downtown . & A man and woman walking down a \textcolor{blue}{busy} street .  \\
    \midrule
    \multicolumn{2}{c}{\textbf{Query: A well dressed man standing in front of \textcolor{blue}{piece of artwork} .}} \\
    \hline
    A well dressed man standing in front of \textcolor{blue}{an abstract fence painting} . & A man wearing headphones is standing in front of a \textcolor{blue}{poster} .  \\
    A man wearing headphones is standing in front of a \textcolor{blue}{poster} . & A man standing in front of a chalkboard points at a \textcolor{blue}{drawing} .      \\
    A man in a blue shirt standing in front of a garage-like structure painted with \textcolor{blue}{geometric designs} .  & A man in a blue shirt standing in front of a garage-like structure painted with \textcolor{blue}{geometric designs} .  \\
    \midrule
    \multicolumn{2}{c}{\textbf{Query: A woman is sitting at a bar \textcolor{blue}{eating a hamburger} .}} \\
    \hline
    A woman sitting \textcolor{blue}{eating a sandwich} . &  A woman is sitting in a cafe \textcolor{blue}{eating lunch} .  \\
    A woman is sitting in a cafe \textcolor{blue}{eating lunch} . &   A woman is \textcolor{blue}{eating at a diner} .  \\
    The woman is \textcolor{blue}{eating a hotdog} in the middle of her bedroom . & A woman is \textcolor{blue}{eating her meal} at a resturant .  \\
    \midrule
    \multicolumn{2}{c}{\textbf{Query: \textcolor{red}{Group of} men trying to \textcolor{blue}{catch fish} with a fishing net .}} \\
    \hline
    \textcolor{red}{Two} men are on a boat trying to \textcolor{blue}{fish for food} during a sunset . & There are \textcolor{red}{three} men on a fishing boat trying to \textcolor{blue}{catch bass} .  \\
    There are \textcolor{red}{three} men on a fishing boat trying to \textcolor{blue}{catch bass} . & \textcolor{red}{Two} men are trying to \textcolor{blue}{fish} .  \\
    \textcolor{red}{Two} men \textcolor{blue}{pull a fishing net up} into their red boat . &  \textcolor{red}{Two} men are on a boat trying to \textcolor{blue}{fish for food} during a sunset .  \\
    \bottomrule[1.2pt]
    \end{tabular}
  \caption{Nearest neighbor retrieval results on the SNLI dataset. Given a a query sentence, the left column shows the top-$3$ retrieved samples based upon the hamming distance with all sentences' binary representations, while the right column exhibits the samples according to the cosine similarity of their continuous embeddings. }
  \vspace{-4mm}
  \label{tab:retrieval}
  \end{scriptsize}
\end{table*}

We also found that our AE-binary-SP model variant consistently demonstrate superior results than the InferLite baselines, which optimize the NLI objective directly over the binary representations. This may be attributed to the difficulty of backpropagating gradients through discrete/binary variables, and would be an interesting direction for future research.
\vspace{-2mm}
\subsection{Nearest Neighbor Retrieval}
\paragraph{Case Study} One major advantage of binary sentence representations is that the similarity of two sentences can be evaluated by merely calculating the hamming distance between their binary codes. To gain more intuition regarding the semantic information encoded in the binary embeddings, we convert all the sentences in the SNLI dataset into continuous and binary vectors (with InferSent-G and AE-binary-SP, respectively). The top-$3$ closet sentences are retrieved based upon the corresponding metrics, and the resulting samples are shown in Table~\ref{tab:retrieval}. It can be observed that the sentences selected based upon the Hamming distance indeed convey very similar semantic meanings. In some cases, the results with binary codes are even more reasonable compared with the continuous embeddings. For example, for the first query, all three sentences in the left column relate to ``watching a movie'', while one of the sentences in the right column is about ``sleeping''.
\vspace{-1mm}
\paragraph{Retrieval Speed} The bitwise comparison is much faster than the element-wise multiplication operation (between real-valued vectors) \citep{Tissier2018NearlosslessBO}. To verify the speed improvement, we sample $10000$ sentence pairs from SNLI and extract their continuous and binary embeddings (with the same dimension of $4096$), respectively. We record the time to compute the cosine similarity and hamming distance between the corresponding representations. With our Python implementation, it takes $3.67\mu$s and $288$ns respectively, indicating that calculating the Hamming distance is over $12$ times faster. Our implementation is not optimized, and the running time of computing Hamming distance can be further improved (to be proportional to the number of different bits, rather than the input length\footnote{\url{https://en.wikipedia.org/wiki/Hamming_distance}}).
\vspace{-1mm}
\subsection{Ablation Study}
\subsubsection{The effect of semantic-preserving loss}
To investigate the importance of incorporating the locality-sensitive regularizer, we select different values of $\lambda_\textit{sp}$ (ranging from $0.0$ to $1.0$) and explore how the transfer results would change accordingly. The $\lambda_\textit{sp}$ controls the relative weight of the semantic-preserving loss term. As shown in Table~\ref{tab:lambda}, augmenting the semantic-preserving loss consistently improves the quality of learned binary embeddings, while the best test accuracy on the MR dataset is obtained with $\lambda_\textit{sp} = 0.8$.
\begin{table} [ht!]
	\centering
	\def\arraystretch{1.2}
	\footnotesize 
	\vspace{-2mm}
	\begin{tabular}{c||c|c|c|c|c}
		\toprule[1.2pt]
		\boldsymbol{$\lambda_\textit{sp}$} & $0.0$ & $0.2$ & $0.5$ & $0.8$ & $1.0$  \\
		\hline
		\textbf{Accuracy} & 78.2 & 78.5 & 78.5  & \tf{79.1} & 78.4  \\
		\bottomrule[1.2pt]
	\end{tabular}
	\caption{Ablation study for the AE-binary-SP model with different choices of $\lambda_\textit{sp}$ (evaluated with test accuracy on the MR dataset).}
	\label{tab:lambda}
	\vspace{-3mm}
\end{table}
\vspace{-1mm}
\subsubsection{Sampling strategy}
\begin{figure} [ht!]
    \centering
     \vspace{-3mm}
    \includegraphics[width = 0.34
    \textwidth]{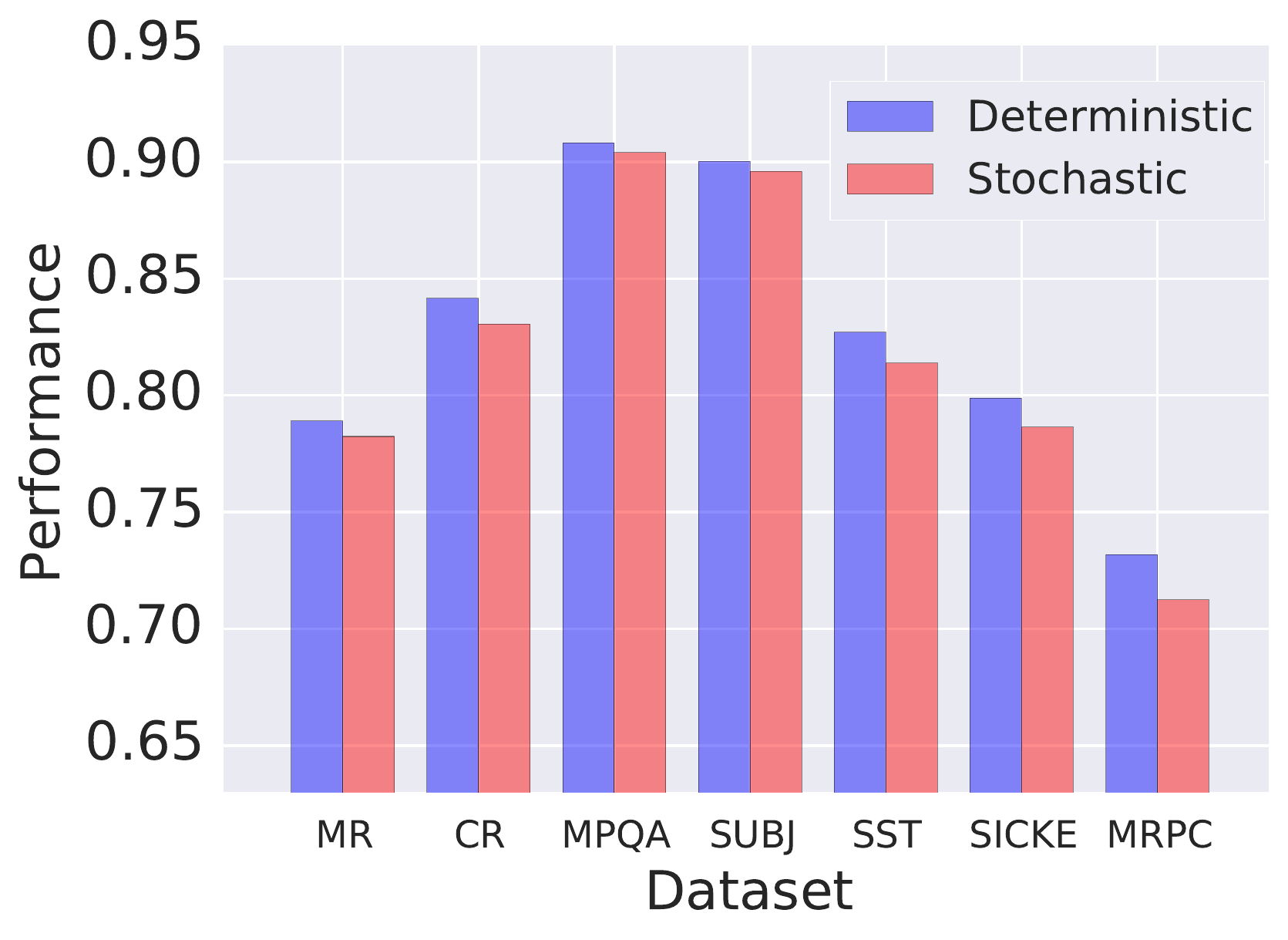}
    \vspace{-3mm}
    \caption{The comparison between \emph{deterministic} and \emph{stochastic} sampling for the autoencoder strategy.}
    \label{fig:deter}
    \vspace{-2mm}
\end{figure}

As discussed in Section~\ref{sec:ae}, the binary latent vector $b$ can be obtained with either a deterministic or stochastically sampled threshold.
We compare these two sampling strategies on several downstream tasks.
As illustrated in Figure~\ref{fig:deter}, setting a fixed threshold demonstrates better empirical performance on all the datasets. Therefore, deterministic threshold is employed for all the autoencoder model variants in our experiments.
\subsubsection{The effect of embedding dimension}
\begin{figure}
    \centering
    \includegraphics[width =0.34\textwidth]{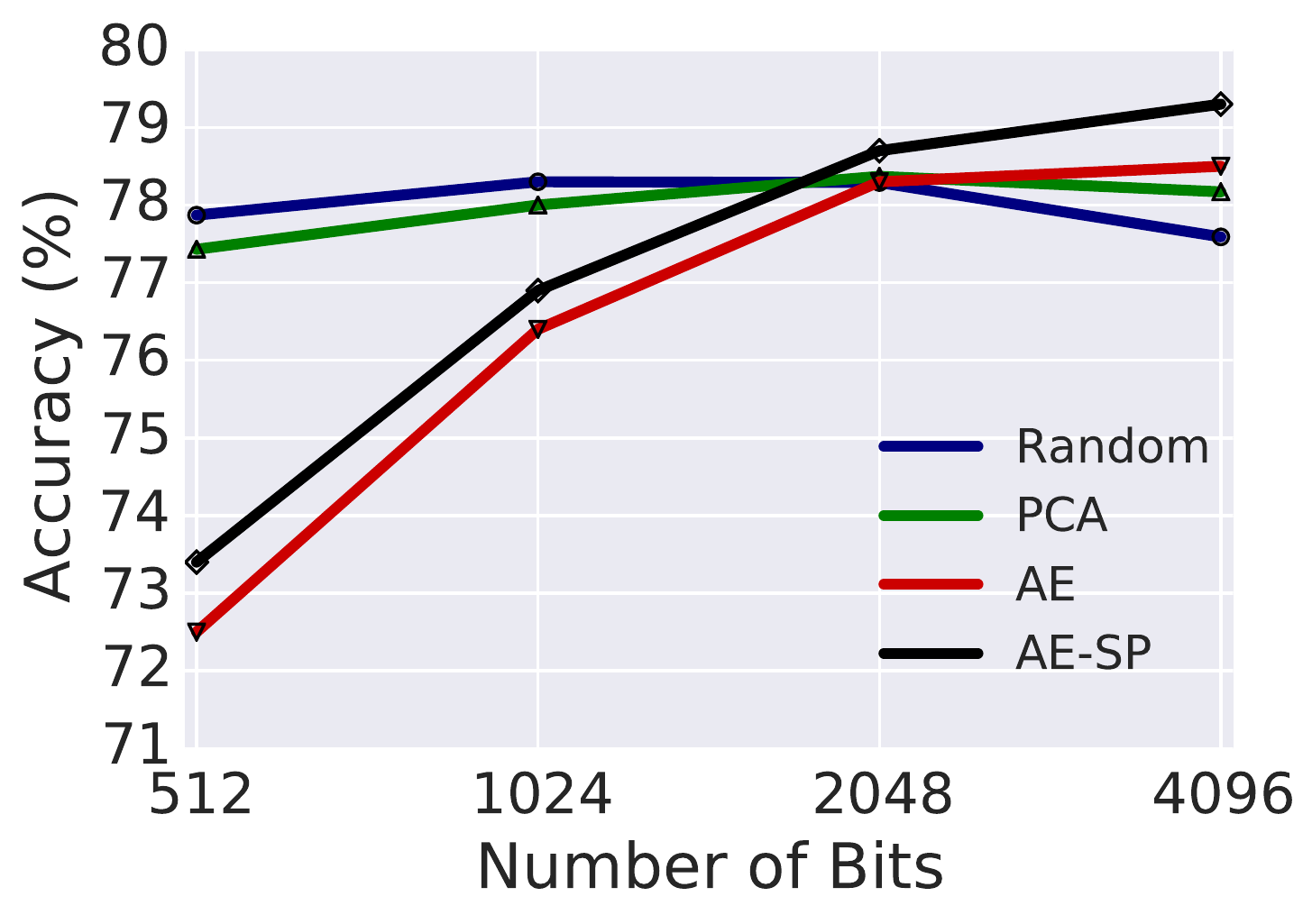}
    \vspace{-3mm}
    \caption{The test accuracy of different model on the MR dataset across $512$,  $1024$,  $2048$,  $4096$ bits for the learned binary representations.}
    \vspace{-3mm}
    \label{fig:dim}
\end{figure}

Except for the hard threshold method, other three proposed strategies all possess the flexibility of adaptively choosing the dimension of learned binary representations. To explore the sensitivity of extracted binary embeddings to their dimensions, we run four model variants (Rand-binary, PCA-binary, AE-binary, AE-binary-SP) with different number of bits (\emph{i.e.}, $512$, $1024$, $2048$, $4096$), and their corresponding results on the MR dataset are shown in Figure~\ref{fig:dim}. 

For the AE-binary and AE-binary-SP models,  
longer binary codes consistently deliver better results. While for the Rand-binary and PCA-binary variants, the quality of inferred representations is much less sensitive to the embedding dimension. Notably, these two strategies exhibit competitive performance even with only $512$ bits. Therefore, in the case where less memory footprint or little training is preferred, Rand-binary and PCA-binary could be more judicious choices. 
\vspace{-2mm}
\section{Conclusion}
\vspace{-2mm}
This paper presents a first step towards learning binary and general-purpose sentence representations that allow for efficient storage and fast retrieval over massive corpora. 
To this end, we explore four distinct strategies to convert pre-trained continuous sentence embeddings into a binarized form.
Notably, a regularized autoencoder augmented with semantic-preserving loss exhibits the best empirical results, degrading performance by only around $2\%$ while saving over $98\%$ memory footprint.
Besides, two other model variants with a random projection or PCA transformation require no training and demonstrate competitive embedding quality even with relatively small dimensions.
Experiments on nearest-neighbor sentence retrieval further validate the effectiveness of proposed framework.

\bibliography{acl2019}

\begin{thebibliography}{39}
\expandafter\ifx\csname natexlab\endcsname\relax\def\natexlab#1{#1}\fi

\bibitem[{Adi et~al.(2017)Adi, Kermany, Belinkov, Lavi, and
  Goldberg}]{Adi2017FinegrainedAO}
Yossi Adi, Einat Kermany, Yonatan Belinkov, Ofer Lavi, and Yoav Goldberg. 2017.
\newblock Fine-grained analysis of sentence embeddings using auxiliary
  prediction tasks.
\newblock \emph{CoRR}, abs/1608.04207.

\bibitem[{Bowman et~al.(2015)Bowman, Angeli, Potts, and
  Manning}]{Bowman2015ALA}
Samuel~R. Bowman, Gabor Angeli, Christopher Potts, and Christopher~D. Manning.
  2015.
\newblock A large annotated corpus for learning natural language inference.
\newblock In \emph{EMNLP}.

\bibitem[{Carreira-Perpin{\'a}n and
  Raziperchikolaei(2015)}]{carreira2015hashing}
Miguel~A Carreira-Perpin{\'a}n and Ramin Raziperchikolaei. 2015.
\newblock Hashing with binary autoencoders.
\newblock In \emph{Proceedings of the IEEE conference on computer vision and
  pattern recognition}, pages 557--566.

\bibitem[{Cer et~al.(2018)Cer, Yang, yi~Kong, Hua, Limtiaco, John, Constant,
  Guajardo-Cespedes, Yuan, Tar, Sung, Strope, and
  Kurzweil}]{Cer2018UniversalSE}
Daniel Cer, Yinfei Yang, Sheng yi~Kong, Nan Hua, Nicole Limtiaco, Rhomni~St.
  John, Noah Constant, Mario Guajardo-Cespedes, Steve Yuan, Chris Tar,
  Yun-Hsuan Sung, Brian Strope, and Ray Kurzweil. 2018.
\newblock Universal sentence encoder.
\newblock \emph{CoRR}, abs/1803.11175.

\bibitem[{Chen et~al.(2018)Chen, Min, and Sun}]{chen2018learning}
Ting Chen, Martin~Renqiang Min, and Yizhou Sun. 2018.
\newblock Learning k-way d-dimensional discrete codes for compact embedding
  representations.
\newblock \emph{arXiv preprint arXiv:1806.09464}.

\bibitem[{Conneau and Kiela(2018)}]{conneau2018senteval}
Alexis Conneau and Douwe Kiela. 2018.
\newblock Senteval: An evaluation toolkit for universal sentence
  representations.
\newblock \emph{arXiv preprint arXiv:1803.05449}.

\bibitem[{Conneau et~al.(2017)Conneau, Kiela, Schwenk, Barrault, and
  Bordes}]{Conneau2017SupervisedLO}
Alexis Conneau, Douwe Kiela, Holger Schwenk, Lo{\"i}c Barrault, and Antoine
  Bordes. 2017.
\newblock Supervised learning of universal sentence representations from
  natural language inference data.
\newblock In \emph{EMNLP}.

\bibitem[{Dai and Le(2015)}]{dai2015semi}
Andrew~M Dai and Quoc~V Le. 2015.
\newblock Semi-supervised sequence learning.
\newblock In \emph{Advances in neural information processing systems}, pages
  3079--3087.

\bibitem[{Dai et~al.(2017)Dai, Guo, Kumar, He, and Song}]{dai2017stochastic}
Bo~Dai, Ruiqi Guo, Sanjiv Kumar, Niao He, and Le~Song. 2017.
\newblock Stochastic generative hashing.
\newblock In \emph{Proceedings of the 34th International Conference on Machine
  Learning-Volume 70}, pages 913--922. JMLR. org.

\bibitem[{Devlin et~al.(2018)Devlin, Chang, Lee, and
  Toutanova}]{devlin2018bert}
Jacob Devlin, Ming-Wei Chang, Kenton Lee, and Kristina Toutanova. 2018.
\newblock Bert: Pre-training of deep bidirectional transformers for language
  understanding.
\newblock \emph{arXiv preprint arXiv:1810.04805}.

\bibitem[{Gan et~al.(2017)Gan, Pu, Henao, Li, He, and
  Carin}]{Gan2017LearningGS}
Zhe Gan, Yunchen Pu, Ricardo Henao, Chunyuan Li, Xiaodong He, and Lawrence
  Carin. 2017.
\newblock Learning generic sentence representations using convolutional neural
  networks.
\newblock In \emph{EMNLP}.

\bibitem[{Glorot and Bengio(2010)}]{glorot2010understanding}
Xavier Glorot and Yoshua Bengio. 2010.
\newblock Understanding the difficulty of training deep feedforward neural
  networks.
\newblock In \emph{Proceedings of the thirteenth international conference on
  artificial intelligence and statistics}, pages 249--256.

\bibitem[{Hill et~al.(2016)Hill, Cho, and Korhonen}]{Hill2016LearningDR}
Felix Hill, Kyunghyun Cho, and Anna Korhonen. 2016.
\newblock Learning distributed representations of sentences from unlabelled
  data.
\newblock In \emph{HLT-NAACL}.

\bibitem[{Hinton(2012)}]{hinton2012neural}
G~Hinton. 2012.
\newblock Neural networks for machine learning. coursera,[video lectures].

\bibitem[{Jang et~al.(2016)Jang, Gu, and Poole}]{jang2016categorical}
Eric Jang, Shixiang Gu, and Ben Poole. 2016.
\newblock Categorical reparameterization with gumbel-softmax.
\newblock \emph{arXiv preprint arXiv:1611.01144}.

\bibitem[{Jernite et~al.(2017)Jernite, Bowman, and
  Sontag}]{Jernite2017DiscourseBasedOF}
Yacine Jernite, Samuel~R. Bowman, and David~A Sontag. 2017.
\newblock Discourse-based objectives for fast unsupervised sentence
  representation learning.
\newblock \emph{CoRR}, abs/1705.00557.

\bibitem[{Joulin et~al.(2016)Joulin, Grave, Bojanowski, and
  Mikolov}]{joulin2016bag}
Armand Joulin, Edouard Grave, Piotr Bojanowski, and Tomas Mikolov. 2016.
\newblock Bag of tricks for efficient text classification.
\newblock \emph{arXiv preprint arXiv:1607.01759}.

\bibitem[{Kingma and Ba(2014)}]{kingma2014adam}
Diederik~P Kingma and Jimmy Ba. 2014.
\newblock Adam: A method for stochastic optimization.
\newblock \emph{arXiv preprint arXiv:1412.6980}.

\bibitem[{Kiros and Chan(2018)}]{kiros2018inferlite}
Jamie Kiros and William Chan. 2018.
\newblock Inferlite: Simple universal sentence representations from natural
  language inference data.
\newblock In \emph{Proceedings of the 2018 Conference on Empirical Methods in
  Natural Language Processing}, pages 4868--4874.

\bibitem[{Kiros et~al.(2015)Kiros, Zhu, Salakhutdinov, Zemel, Torralba,
  Urtasun, and Fidler}]{Kiros2015SkipThoughtV}
Ryan Kiros, Yukun Zhu, Ruslan Salakhutdinov, Richard~S. Zemel, Antonio
  Torralba, Raquel Urtasun, and Sanja Fidler. 2015.
\newblock Skip-thought vectors.
\newblock In \emph{NIPS}.

\bibitem[{Logeswaran and Lee(2018)}]{Logeswaran2018AnEF}
Lajanugen Logeswaran and Honglak Lee. 2018.
\newblock An efficient framework for learning sentence representations.
\newblock \emph{ICLR}.

\bibitem[{Nie et~al.(2017)Nie, Bennett, and Goodman}]{Nie2017DisSentSR}
Allen Nie, Erin~D. Bennett, and Noah~D. Goodman. 2017.
\newblock Dissent: Sentence representation learning from explicit discourse
  relations.
\newblock \emph{CoRR}, abs/1710.04334.

\bibitem[{Pagliardini et~al.(2018)Pagliardini, Gupta, and
  Jaggi}]{Pagliardini2018UnsupervisedLO}
Matteo Pagliardini, Prakhar Gupta, and Martin Jaggi. 2018.
\newblock Unsupervised learning of sentence embeddings using compositional
  n-gram features.
\newblock In \emph{NAACL-HLT}.

\bibitem[{Radford et~al.(2018)Radford, Narasimhan, Salimans, and
  Sutskever}]{radford2018improving}
Alec Radford, Karthik Narasimhan, Tim Salimans, and Ilya Sutskever. 2018.
\newblock Improving language understanding by generative pre-training.
\newblock \emph{URL https://s3-us-west-2. amazonaws.
  com/openai-assets/research-covers/languageunsupervised/language understanding
  paper. pdf}.

\bibitem[{Ravi and Kozareva(2018)}]{ravi2018self}
Sujith Ravi and Zornitsa Kozareva. 2018.
\newblock Self-governing neural networks for on-device short text
  classification.
\newblock In \emph{Proceedings of the 2018 Conference on Empirical Methods in
  Natural Language Processing}, pages 804--810.

\bibitem[{Ruder and Howard(2018)}]{Ruder2018UniversalLM}
Sebastian Ruder and Jeremy Howard. 2018.
\newblock Universal language model fine-tuning for text classification.
\newblock In \emph{ACL}.

\bibitem[{Salakhutdinov and Hinton(2009)}]{salakhutdinov2009semantic}
Ruslan Salakhutdinov and Geoffrey Hinton. 2009.
\newblock Semantic hashing.
\newblock \emph{International Journal of Approximate Reasoning},
  50(7):969--978.

\bibitem[{Shen et~al.(2018)Shen, Su, Chapfuwa, Wang, Wang, Carin, and
  Henao}]{Shen2018NASHTE}
Dinghan Shen, Qinliang Su, Paidamoyo Chapfuwa, Wenlin Wang, Guoyin Wang,
  Lawrence Carin, and Ricardo Henao. 2018.
\newblock Nash: Toward end-to-end neural architecture for generative semantic
  hashing.
\newblock In \emph{ACL}.

\bibitem[{Shu and Nakayama(2017)}]{shu2017compressing}
Raphael Shu and Hideki Nakayama. 2017.
\newblock Compressing word embeddings via deep compositional code learning.
\newblock \emph{arXiv preprint arXiv:1711.01068}.

\bibitem[{Tang and de~Sa(2018)}]{tang2018improving}
Shuai Tang and Virginia~R de~Sa. 2018.
\newblock Improving sentence representations with multi-view frameworks.
\newblock \emph{arXiv preprint arXiv:1810.01064}.

\bibitem[{Tissier et~al.(2019)Tissier, Habrard, and
  Gravier}]{Tissier2018NearlosslessBO}
Julien Tissier, Amaury Habrard, and Christophe Gravier. 2019.
\newblock Near-lossless binarization of word embeddings.
\newblock \emph{AAAI}.

\bibitem[{Van~Durme and Lall(2010)}]{van2010online}
Benjamin Van~Durme and Ashwin Lall. 2010.
\newblock Online generation of locality sensitive hash signatures.
\newblock In \emph{Proceedings of the ACL 2010 Conference Short Papers}, pages
  231--235. Association for Computational Linguistics.

\bibitem[{Wang et~al.(2014)Wang, Shen, Song, and Ji}]{wang2014hashing}
Jingdong Wang, Heng~Tao Shen, Jingkuan Song, and Jianqiu Ji. 2014.
\newblock Hashing for similarity search: A survey.
\newblock \emph{arXiv preprint arXiv:1408.2927}.

\bibitem[{Wieting et~al.(2016)Wieting, Bansal, Gimpel, and
  Livescu}]{Wieting2016TowardsUP}
John Wieting, Mohit Bansal, Kevin Gimpel, and Karen Livescu. 2016.
\newblock Towards universal paraphrastic sentence embeddings.
\newblock \emph{CoRR}, abs/1511.08198.

\bibitem[{Wieting and Gimpel(2018)}]{Wieting2018ParaNMT50MPT}
John Wieting and Kevin Gimpel. 2018.
\newblock Paranmt-50m: Pushing the limits of paraphrastic sentence embeddings
  with millions of machine translations.
\newblock In \emph{ACL}.

\bibitem[{Wieting and Kiela(2018)}]{Wieting2018NoTR}
John Wieting and Douwe Kiela. 2018.
\newblock No training required: Exploring random encoders for sentence
  classification.
\newblock \emph{CoRR}, abs/1901.10444.

\bibitem[{Williams et~al.(2017)Williams, Nangia, and
  Bowman}]{williams2017broad}
Adina Williams, Nikita Nangia, and Samuel~R Bowman. 2017.
\newblock A broad-coverage challenge corpus for sentence understanding through
  inference.
\newblock \emph{arXiv preprint arXiv:1704.05426}.

\bibitem[{Xu et~al.(2015)Xu, Wang, Tian, Xu, Zhao, Wang, and
  Hao}]{xu2015convolutional}
Jiaming Xu, Peng Wang, Guanhua Tian, Bo~Xu, Jun Zhao, Fangyuan Wang, and
  Hongwei Hao. 2015.
\newblock Convolutional neural networks for text hashing.
\newblock In \emph{Twenty-Fourth International Joint Conference on Artificial
  Intelligence}.

\bibitem[{Zhang et~al.(2010)Zhang, Wang, Cai, and Lu}]{zhang2010self}
Dell Zhang, Jun Wang, Deng Cai, and Jinsong Lu. 2010.
\newblock Self-taught hashing for fast similarity search.
\newblock In \emph{Proceedings of the 33rd international ACM SIGIR conference
  on Research and development in information retrieval}, pages 18--25. ACM.

\end{thebibliography}
\bibliographystyle{acl_natbib}

\end{document}